\documentclass[letterpaper, 10 pt, conference]{ieeeconf}  

\IEEEoverridecommandlockouts                              

\usepackage{amsmath,amssymb}
\usepackage{graphicx}
\usepackage{cite}

\title{\LARGE \bf
q-VAE for Disentangled Representation Learning \\and Latent Dynamical Systems
}

\author{Taisuke Kobayashi$^{1}$%
\thanks{$^{1}$Taisuke Kobayashi is with the Division of Information Science, Nara Institute of Science and Technology, 8916-5 Takayama-cho, Ikoma, Nara 630-0192, Japan
        {\tt\footnotesize kobayashi@is.naist.jp}}%
}

\begin{document}

\maketitle
\thispagestyle{empty}
\pagestyle{empty}

\begin{abstract}

A variational autoencoder (VAE) derived from Tsallis statistics called q-VAE is proposed.
In the proposed method, a standard VAE is employed to statistically extract latent space hidden in sampled data, and this latent space helps make robots controllable in feasible computational time and cost.
To improve the usefulness of the latent space, this paper focuses on disentangled representation learning, e.g., $\beta$-VAE, which is the baseline for it.
Starting from a Tsallis statistics perspective, a new lower bound for the proposed q-VAE is derived to maximize the likelihood of the sampled data, which can be considered an adaptive $\beta$-VAE with deformed Kullback-Leibler divergence.
To verify the benefits of the proposed q-VAE, a benchmark task to extract the latent space from the MNIST dataset was performed.
The results demonstrate that the proposed q-VAE improved disentangled representation while maintaining the reconstruction accuracy of the data.
In addition, it relaxes the independency condition between data, which is demonstrated by learning the latent dynamics of nonlinear dynamical systems.
By combining disentangled representation, the proposed q-VAE achieves stable and accurate long-term state prediction from the initial state and the action sequence.

\end{abstract}

\section{Introduction}

Recently, deep learning~\cite{lecun2015deep} has become the most powerful tool to resolve analytically unsolvable problems.
In robotics and robot control, several studies have applied deep learning to perform complicated tasks that rely on raw images from cameras~\cite{kalashnikov2018scalable,tsurumine2019deep}.
Typically, deep learning requires a large number of samples for end-to-end learning, i.e., from extracting features hidden in the inputs (e.g., images) to optimal control based on the extracted features.

Modularizing various functions, e.g., feature extraction~\cite{kingma2014auto,kobayashi2015selection} and optimal control~\cite{neunert2016fast,kobayashi2018unified,koenemann2015whole,amos2018differentiable}, is effective at reducing the number of samples.
In particular, control theory has long been studied relative to stability, convergence speed, etc.
Therefore, exploiting conventional but powerful control technology is a solution to reducing the number of samples; however, features extracted without ingenuity would be unsuitable.

Thus, this paper focuses on methods to extract features hidden in inputs.
To this end, the variational autoencoder (VAE)~\cite{kingma2014auto} is a promising method. The VAE encodes inputs to a latent space with stochastic latent variables (and decodes them to inputs) in an unsupervised manner.
In fact, control methods for the latent space gained by VAE have been proposed previously~\cite{watter2015embed,fraccaro2017disentangled,hafner2019learning}.

To improve the usefulness of such a latent space, recent VAE research has investigated disentangled representation learning~\cite{higgins2018towards,van2019disentangled}, which assigns independent attributes hidden in the inputs to the axes of the latent space without supervisory signals.
However, a representative of this methodology, i.e., $\beta$-VAE~\cite{higgins2017beta}, would lose the value of the VAE as a data generation model because the reconstruction accuracy of the inputs is reduced.
Although variants of $\beta$-VAE have been proposed to resolve this problem, they have some drawbacks from a practicality perspective, e.g., heuristic designs are difficult to optimize~\cite{burgess2018understanding}, the assumption of batch data~\cite{chen2018isolating}, and optimization of an additional discriminator network~\cite{kim2018disentangling}.

To realize simple and low-cost disentangled representation learning that is practically sufficient, this paper proposes a derivation of VAE combined with Tsallis statistics~\cite{tsallis1988possible,suyari2005law,umarov2008q,nielsen2011closed}, which refer to the extended version of general statistics based on real parameter $q$.
This derivation, which we refer to as q-VAE, mathematically provides adaptive $\beta$ according to the amount of information in latent variables.
Due to the adaptive $\beta$, the proposed q-VAE achieves proper extraction of the essential information of inputs while maintaining the reconstruction accuracy of the inputs.
In addition, deformed Kullback-Leibler (KL) divergence~\cite{nielsen2011closed,gil2013renyi} weakens the magnitude of regularization in the vicinity of the center location of prior, which may construct the meaningful latent space.

In addition, the proposed q-VAE relaxes the condition that inputs be independent and identically distributed (i.i.d.)~\cite{umarov2008q}, which, generally, cannot be satisfied in a functional manner when controlling robots.
Therefore, q-VAE is also extended to a model to learn the latent dynamics of inputs when manipulated variables are given.
As a result, the encoded latent variables are transited to the next ones that are decoded to the next inputs.
With a disentangled representation, even if the latent dynamics model is constrained as a diagonal system (i.e., a model in which latent variables are independent), which could be easily exploited to control robots with low computational cost, the proposed q-VAE alleviates the deterioration of prediction performance.

We verified the proposed q-VAE using an MNIST benchmark. The results indicate that q-VAE with an appropriate parameter outperforms $\beta$-VAE (i.e., fixed $\beta$) in terms of disentangled representation and the reconstruction accuracy of inputs.
In addition, learning the latent dynamics in nonlinear systems was performed.
Although conventional methods fail to predict future states in a stable manner, q-VAE realizes stable and accurate long-term state prediction from the initial state and the action sequence.

\section{Preliminaries}

\subsection{Variational autoencoder}


In this section, we briefly introduce the VAE~\cite{kingma2014auto} (see the upper structure of Fig.~\ref{fig:structure_dynamics}).
Here, a generative model of inputs $\boldsymbol{x}$ from latent variables $\boldsymbol{z}$ is considered.
This decoder is approximated by (deep) neural networks with parameters $\boldsymbol{\theta}$: $p(\boldsymbol{x} \mid \boldsymbol{z}; \boldsymbol{\theta})$.
The VAE attempts to maximize the log likelihood of $N$ inputs, $\log{p(X)}$, where $X = \{\boldsymbol{x}_n\}_1^N$.
By assuming that the inputs are i.i.d., $\log{p(X)}$ can be simplified to the sum of the respective log likelihoods $\sum_{n=1}^N \log{\boldsymbol{x}_n}$.
To maximize $\log{p(X)}$ indirectly, an evidence lower bound (ELBO) $\mathcal{L}(X)$, which is derived using Jensen's inequality and an encoder with parameters $\boldsymbol{\phi}$, $\rho(\boldsymbol{z} \mid \boldsymbol{x}; \boldsymbol{\phi})$, is maximized.
Then, $\mathcal{L}(X)$ is derived as follows.
\begin{align}
    \log{p(X)} &= \sum_{n=1}^N \log \int p(\boldsymbol{x}_n \mid \boldsymbol{z} ; \boldsymbol{\theta}) p(\boldsymbol{z}) d\boldsymbol{z}
    \nonumber\\
    &= \sum_{n=1}^N \log \int \frac{\rho(\boldsymbol{z} \mid \boldsymbol{x}_n ; \boldsymbol{\phi})}{\rho(\boldsymbol{z} \mid \boldsymbol{x}_n ; \boldsymbol{\phi})} p(\boldsymbol{x}_n \mid \boldsymbol{z} ; \boldsymbol{\theta}) p(\boldsymbol{z}) d\boldsymbol{z}
    \nonumber\\
    &\geq \sum_{n=1}^N \mathbb{E}_{\rho(\boldsymbol{z} \mid \boldsymbol{x}_n ; \boldsymbol{\phi})}\left [ \log{\frac{p(\boldsymbol{x}_n \mid \boldsymbol{z} ; \boldsymbol{\theta}) p(\boldsymbol{z})}{\rho(\boldsymbol{z} \mid \boldsymbol{x}_n ; \boldsymbol{\phi})}} \right ]
    \nonumber\\
    &=: \mathcal{L}(X)
    \label{eq:derive_elbo_vae}
\end{align}
where $p(\boldsymbol{z})$ is a prior of the latent variables.
$\mathcal{L}(X)$ can be expressed as follows.
\begin{align}
    \mathcal{L}(X) &= \sum_{n=1}^N \mathbb{E}_{\rho(\boldsymbol{z} \mid \boldsymbol{x}_n ; \boldsymbol{\phi})}[\log{p(\boldsymbol{x}_n \mid \boldsymbol{z} ; \boldsymbol{\theta})}]
    \nonumber\\
    &- \mathrm{KL}(\rho(\boldsymbol{z} \mid \boldsymbol{x}_n ; \boldsymbol{\phi}) \mid \mid p(\boldsymbol{z}))
    \nonumber\\
    &\simeq \sum_{n=1}^N \log{p(\boldsymbol{x}_n \mid \boldsymbol{z}_n ; \boldsymbol{\theta})} - \mathrm{KL}(\rho(\boldsymbol{z} \mid \boldsymbol{x}_n ; \boldsymbol{\phi}) \mid \mid p(\boldsymbol{z}))
    \label{eq:elbo_vae}
\end{align}
where the first term denotes the negative reconstruction error in the standard autoencoder, and the second term, i.e., the KL divergence between the posterior and prior, is the regularization term, which is used to attempt to $\boldsymbol{z} \sim p(\boldsymbol{z})$.
Generally, $p(\boldsymbol{z})$ is given as standard normal distribution $\mathcal{N}(\boldsymbol{0}, I)$.

In the case of $\beta$-VAE~\cite{higgins2017beta}, the second term is multiplied by $\beta \in (0, \infty)$, although its derivation is given by solving a constrained optimization problem.
By giving $\beta > 1$, regularization to $p(\boldsymbol{z})$ is strengthened, and the role of each axis of the latent space, which has limited expressive ability, is clarified to reconstruct the inputs.

\subsection{Tsallis statistics}

\begin{figure}[tb]
    \centering
    \includegraphics[keepaspectratio=true,width=0.72\linewidth]{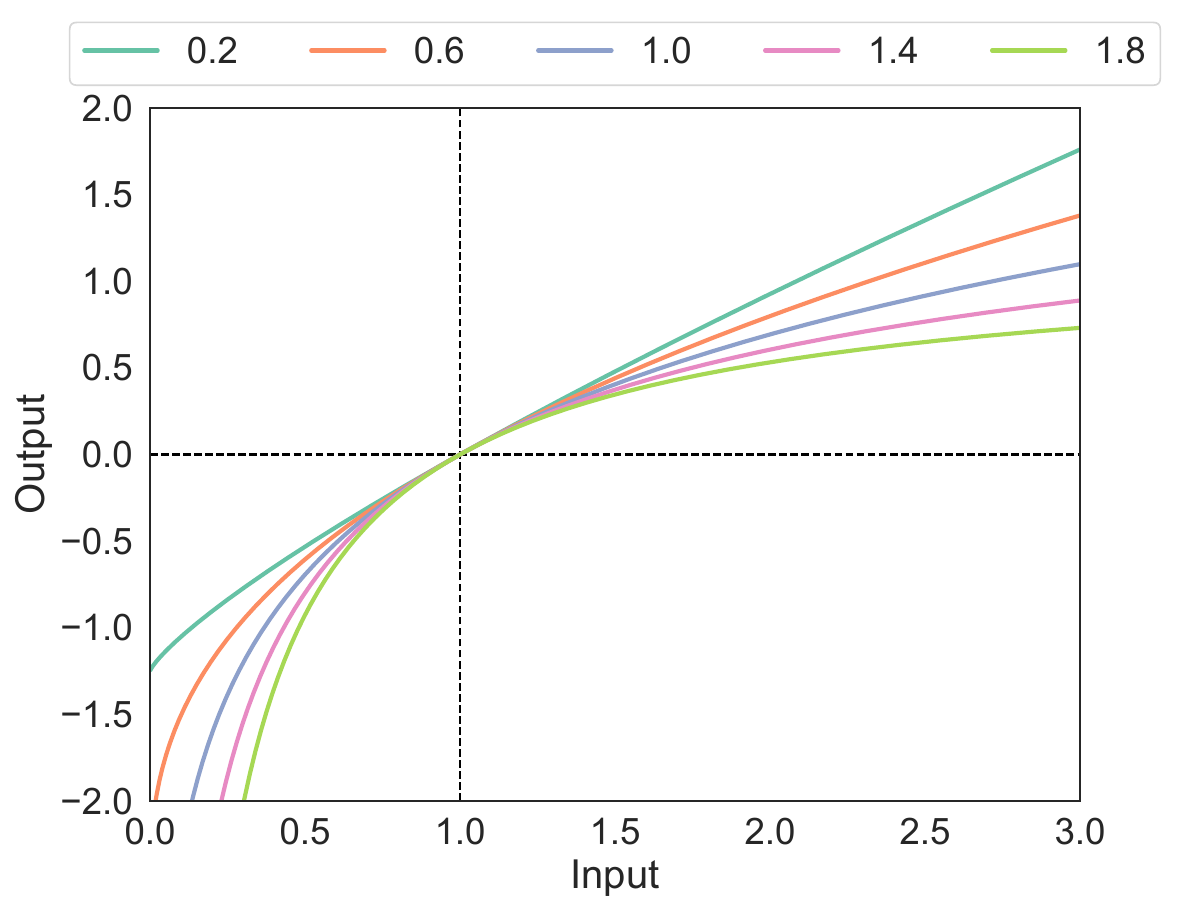}
    \caption{Examples of $q$-logarithm with different $q$ values.
        When $q > 0$, this function is considered as a concave function.
        As can be seen in the intuitive, $\ln_{q+\epsilon}(x)$ with $\epsilon$ a positive small number is larger than $\ln_q(x)$.
    }
    \label{fig:plot_lnq}
\end{figure}

Tsallis statistics refers to the organization of mathematical functions and associated probability distributions proposed by Tsallis~\cite{tsallis1988possible}.
This concept is organized based on $q$-deformed exponential and logarithmic functions, which are extensions of general exponential and logarithmic functions by real number $q \in \mathbb{R}$.
We introduce the following definitions From Tsallis statistics.

First, the $q$-logarithm, $\ln_q(x)$ with $x > 0$, is given as follows.
\begin{align}
    \ln_q(x) =
    \begin{cases}
        \ln(x) & q = 1
        \\
        \frac{x^{1 - q} - 1}{1 - q} & q \neq 1
        \\
    \end{cases}
    \label{eq:q_log}
\end{align}
where $q$ gives its shape, as shown in Fig.~\ref{fig:plot_lnq}.
As shown in Fig.~\ref{fig:plot_lnq}, the $q$-logarithm with $q > 0$ is a concave function.
Note that when $q \to 1$, the lower equation converges to the natural logarithm.

In the $q$-logarithm, multiplication of two variables, $x, y > 0$, is no longer simply divided.
In other words, the following pseudo-additivity is derived.
\begin{align}
    \ln_q(xy) = \ln_q(x) + \ln_q(y) + (1 - q) \ln_q(x) \ln_q(y)
    \label{eq:p_add}
\end{align}

Rather than general multiplication, a new multiplication operation $\otimes_q$ is introduced as follows~\cite{suyari2005law}.
\begin{align}
    x \otimes_q y =
    \begin{cases}
        (x^{1-q} + y^{1-q} - 1)^\frac{1}{1 - q} & x^{1-q} + y^{1-q} > 1
        \\
        0 & \mathrm{otherwise}
        \\
    \end{cases}
    \label{eq:q_mul}
\end{align}
This definition means that the following additivity is satisfied when using $\otimes_q$.
\begin{align}
    \ln_q(x \otimes_q y) = \ln_q(x) + \ln_q(y)
    \label{eq:q_add}
\end{align}
By using $\otimes_q$, the $q$-likelihood of data $X = \{ \boldsymbol{x}_n\}_1^N$, which is maximized when the probability is given as $q$-Gaussian, can be defined with q-i.i.d. (a relaxed version of the i.i.d. condition~\cite{suyari2005law,umarov2008q}).
\begin{align}
    p(X) = p(\boldsymbol{x}_1) \otimes_q p(\boldsymbol{x}_2) \otimes_q \cdots \otimes_q p(\boldsymbol{x}_N)
    \label{eq:q_likelihood}
\end{align}

Finally, the deformed version of KL divergence (referred to as Tsallis divergence~\cite{nielsen2011closed}), $\mathrm{KL}_q$, is expressed as follows.
\begin{align}
    \mathrm{KL}_q(p_1 \mid \mid p_2) = - \int p_1(x) \ln_q\frac{p_2(x)}{p_1(x)} dx
    \label{eq:q_kld}
\end{align}
where $p_1$ and $p_2$ are arbitrary probability density functions.
The $q$-logarithm for all $x$ decreases as $q$ increases; thus, $\mathrm{KL}_q$ increases as $q$ increases.
Tsallis divergence can be derived by transforming Renyi divergence, which has closed-form solutions for commonly-used distributions~\cite{gil2013renyi}.
Therefore, the proposed q-VAE can be integrated with different priors, e.g., the Laplace distribution and mixture models.

\section{q-variational autoencoder }

\subsection{Derivation of ELBO}

By combining Eqs.~\eqref{eq:q_add} and~\eqref{eq:q_likelihood}, a new $q$-log likelihood to be maximized is derived.
In addition, if $q > 0$, the $q$-logarithm is a concave function (Fig.~\ref{fig:plot_lnq}); therefore, Jensen's inequality can be used in a manner similar to Eq.~\eqref{eq:derive_elbo_vae}.

\begin{align}
    \ln_q p(X) &= \sum_{n=1}^N \ln_q \int p(\boldsymbol{x}_n \mid \boldsymbol{z} ; \boldsymbol{\theta}) p(\boldsymbol{z}) d\boldsymbol{z}
    \nonumber\\
    &\geq \sum_{n=1}^N \mathbb{E}_{\rho(\boldsymbol{z} \mid \boldsymbol{x}_n ; \boldsymbol{\phi})}\left [ \ln_q{\frac{p(\boldsymbol{x}_n \mid \boldsymbol{z} ; \boldsymbol{\theta}) p(\boldsymbol{z})}{\rho(\boldsymbol{z} \mid \boldsymbol{x}_n ; \boldsymbol{\phi})}} \right ]
    \nonumber\\
    &=: \mathcal{L}_q(X)
    \label{eq:derive_elbo_qvae}
\end{align}
Note again that $X$ can be relaxed to be q-i.i.d. data, unlike the log likelihood used in Eq.~\eqref{eq:derive_elbo_vae}.

According to Eq.~\eqref{eq:p_add}, $\mathcal{L}_q(X)$ can be divided into three terms and summarized to two terms similar to Eq.~\eqref{eq:elbo_vae}.
\begin{align}
    \mathcal{L}_q(X) &= \sum_{n=1}^N \mathbb{E}_{\rho(\boldsymbol{z} \mid \boldsymbol{x}_n ; \boldsymbol{\phi})}\bigg[
    \ln_q{p(\boldsymbol{x}_n \mid \boldsymbol{z} ; \boldsymbol{\theta})}
    \nonumber\\
    &+ \ln_q{\frac{p(\boldsymbol{z})}{\rho(\boldsymbol{z} \mid \boldsymbol{x}_n ; \boldsymbol{\phi})}}
    \nonumber\\
    &+ (1 - q) \ln_q{p(\boldsymbol{x}_n \mid \boldsymbol{z} ; \boldsymbol{\theta})} \ln_q{\frac{p(\boldsymbol{z})}{\rho(\boldsymbol{z} \mid \boldsymbol{x}_n ; \boldsymbol{\phi})}} \bigg]
    \nonumber\\
    &\simeq \sum_{n=1}^N
    \ln_q{p(\boldsymbol{x}_n \! \mid \! \boldsymbol{z}_n ; \! \boldsymbol{\theta})}
    \! \left \{ \! 1 \! + \! (1 \! - \! q) \ln_q{\frac{p(\boldsymbol{z}_n)}{\rho(\boldsymbol{z}_n \! \mid \! \boldsymbol{x}_n ; \! \boldsymbol{\phi})}} \right \}
    \nonumber\\
    & - \mathrm{KL}_q(\rho(\boldsymbol{z} \mid \boldsymbol{x}_n ; \boldsymbol{\phi}) \mid \mid p(\boldsymbol{z}))
    \nonumber\\
    &= \sum_{n=1}^N
    \frac{\ln_q{p(\boldsymbol{x}_n \mid \boldsymbol{z}_n ; \boldsymbol{\theta})}}{\beta_q(\boldsymbol{x}_n, \boldsymbol{z}_n)} - \mathrm{KL}_q(\rho(\boldsymbol{z} \mid \boldsymbol{x}_n ; \boldsymbol{\phi}) \mid \mid p(\boldsymbol{z}))
    \label{eq:elbo_qvae}
\end{align}
where the first term is denotes the negative reconstruction error with a new adaptive variable $1/\beta_q(\boldsymbol{x}_n, \boldsymbol{z}_n)$, and the second term, i.e., Tsallis divergence between the posterior and prior, is the regularization term used to attempt to $\boldsymbol{z} \sim p(\boldsymbol{z})$.

\subsection{Practical implementation}
\label{subsec:implementation}

Here, we present four practical statements.
First, the probability output from decoder $p(\boldsymbol{x} \mid \boldsymbol{z} ; \boldsymbol{\theta})$ should be assumed as q-Gaussian distribution (or a Bernoulli distribution for binary inputs) to match the reconstruction error term with that of the standard VAE.
Although a Gaussian distribution can be assumed, it would be numerically unstable because its exponential function is not canceled by logarithm and the power function is added by the q-logarithm.
In addition, q-Gaussian distribution includes the student-t distribution, which yields robust estimation~\cite{takahashi2018student}; therefore, further investigation of the decoder model may further unlock the potential of q-VAE.

Second, the computational graph of $\beta_q(\boldsymbol{x}, \boldsymbol{z})$ is cut to simplify backpropagation and regard it as simply the input-dependent coefficient.
Even with the computational graph, the learning direction of parameters (see the next section) would be the same as the case without it.
However, in the case with the computational graph, the gradient scale varies, thereby making learning unstable if the same hyperparameters as the standard VAE are employed.
In fact, that case happened learning failure in debugging.

Third, we employ the latent distribution as Gaussian, which has a closed-form solution of Tsallis divergence ~\cite{nielsen2011closed,gil2013renyi}.
Given $p_1$ and $p_2$ as $d$-dimensional normal distributions with parameters $\boldsymbol{\mu}_1$, $\Sigma_1$, $\boldsymbol{\mu}_2$, and $\Sigma_2$, Tsallis divergence is solved as follows.
\begin{align}
    \mathrm{KL}_q(p_1 \mid \mid p_2) =
    \begin{cases}
        \frac{1}{2} \left \{ \mathrm{tr}(\Sigma_2^{-1} \Sigma_1) + \ln \frac{|\Sigma_2|}{|\Sigma_1|}  - d \right . &
        \\
        \left . + (\boldsymbol{\mu_2} - \boldsymbol{\mu_1})^\top \Sigma_2^{-1} (\boldsymbol{\mu_2} - \boldsymbol{\mu_1}) \right \} & q = 1
        \\
        \frac{\exp \left ( \frac{1}{2}I_q(p_1 \mid \mid p_2) \right ) - 1}{q - 1} & q \neq 1
        \\
    \end{cases}
    \label{eq:q_kld_normal}
\end{align}
where,
\begin{align}
    I_q(p_1 \mid \mid p_2) &= \ln \frac{|\Sigma_2|^q |\Sigma_1|^{1-q}}{|\Sigma|}
    \nonumber\\
    &+ q (1-q) (\boldsymbol{\mu_2} - \boldsymbol{\mu_1})^\top \Sigma^{-1} (\boldsymbol{\mu_2} - \boldsymbol{\mu_1})
    \\
    \Sigma &= q \Sigma_2 + (1-q) \Sigma_1
    \label{eq:q_kld_normal_cov}
\end{align}

Finally, according to the derivation of $\beta$-VAE~\cite{higgins2017beta}, q-VAE can be integrated with $\beta$-VAE.
In other words, the following constrained optimization problem is solved.
\begin{align}
    & \max_{\boldsymbol{\theta}, \boldsymbol{\phi}} \sum_{n=1}^N
    \frac{\ln_q{p(\boldsymbol{x}_n \mid \boldsymbol{z}_n ; \boldsymbol{\theta})}}{\beta_q(\boldsymbol{x}_n, \boldsymbol{z}_n)}
    \nonumber\\
    \mathrm{s.t.}\  & \mathrm{KL}_q(\rho(\boldsymbol{z} \mid \boldsymbol{x}_n ; \boldsymbol{\phi}) \mid \mid p(\boldsymbol{z})) < \epsilon
    \nonumber
\end{align}
where $\epsilon$ denotes a threshold of Tsallis divergence.
This problem can be rewritten as a Lagrangian under the KKT conditions as follows.
\begin{align}
    \mathcal{L}_{(\beta, q)}(X) &= \sum_{n=1}^N
    \frac{\ln_q{p(\boldsymbol{x}_n \mid \boldsymbol{z}_n ; \boldsymbol{\theta})}}{\beta_q(\boldsymbol{x}_n, \boldsymbol{z}_n)}
    \nonumber\\
    &- \beta \mathrm{KL}_q(\rho(\boldsymbol{z} \mid \boldsymbol{x}_n ; \boldsymbol{\phi}) \mid \mid p(\boldsymbol{z}))
    \label{eq:elbo_beta_qvae}
\end{align}
where $\beta$ denotes the hyperparameter to tune the tradeoff between reconstruction and regularization.

\subsection{Analysis}

In the proposed q-VAE, the above ELBO is maximized by optimizing parameter $\boldsymbol{\theta}$ of the decoder and parameter $\boldsymbol{\phi}$ of the encoder.
As shown in Eqs.~\eqref{eq:elbo_vae} and \eqref{eq:elbo_qvae}, q-VAE extends the standard VAE by adding parameter $q$ because, if $q \to 1$, $\ln_q$, $\beta_q$, and $\mathrm{KL}_q$ converge on $\ln$, $1$, and $\mathrm{KL}$, respectively.

In addition, the proposed q-VAE can be considered a type of $\beta$-VAE with adaptive $\beta_q(\boldsymbol{x}_n, \boldsymbol{z}_n)$.
According to range of $q$, the following three cases are expected.
\begin{enumerate}
    \item $q=1$: As mentioned previously, $\beta_q$ is always equal to $1$ in this case.
    Therefore, this case is considered the standard VAE or $\beta$-VAE with hyperparameter $\beta$.
    \item $q<1$: Generally, the posterior $\rho(\boldsymbol{z}_n \mid \boldsymbol{x}_n ; \boldsymbol{\phi})$ contains more information than the prior $p(\boldsymbol{z}_n)$.
    Thus, $\beta_q$ would likely be greater than $1$.
    In other words, the posterior is strongly constrained to discard information about inputs only when it has a significant amount of information. Consequently, only essential information of the inputs is expected to be left without duplication in the latent space (and its axes).
    However, smaller $q$ leads to smaller $\mathrm{KL}_q$; thus, if $q$ is too small, no matter how large $\beta_q$ is, the constraint to the prior will not work as described above because it was originally small.
    \item $q>1$: In contrast to the $q<1$ case, $\beta_q$ would likely be less than $1$.
    In other words, reconstruction would be prioritized without the constraint to the prior.
    In addition, as shown in Eq.~\eqref{eq:q_kld_normal_cov}, the covariance matrix for the Tsallis divergence calculation is likely to violate its positive-semidefinite condition.
\end{enumerate}

From the above, we expect that a $q$ value that is less than $1$ is better relative to achieving disentangled representation learning with stable computation.
To the best of our knowledge, no methods have been proposed to date to automatically tune $\beta$ in $\beta$-VAE.
Therefore, the proposed q-VAE method is the first method that can tune $\beta$ via a mathematically natural derivation.

To avoid adverse effects caused by excessively small $q$ values, the simplified version of Eq.~\eqref{eq:elbo_beta_qvae} is investigated in the following.
\begin{align}
    \mathcal{L}_{(\beta, q)}^\mathrm{smp}(X) &= \sum_{n=1}^N
    \frac{\ln_q{p(\boldsymbol{x}_n \mid \boldsymbol{z}_n ; \boldsymbol{\theta})}}{\beta_q(\boldsymbol{x}_n, \boldsymbol{z}_n)}
    \nonumber\\
    &- \beta \mathrm{KL}(\rho(\boldsymbol{z} \mid \boldsymbol{x}_n ; \boldsymbol{\phi}) \mid \mid p(\boldsymbol{z}))
    \label{eq:elbo_beta_qvae2}
\end{align}
This means that $\mathrm{KL}$ (rather than $\mathrm{KL}_q$) is employed for the strong constraint.
When $q < 1$, $\mathcal{L}_{(\beta, q)}^\mathrm{smp}$ is less than the original $\mathcal{L}_{(\beta, q)}$ because $\mathrm{KL}$ is stronger than $\mathrm{KL}_q$.
Even though this case would avoid an overly weak constraint by $\mathrm{KL}_q$ with $q \ll 1$, it may make the constraint stronger when $q \simeq 1$.

\subsection{Extension to latent dynamical systems}

\begin{figure}[tb]
    \centering
    \includegraphics[keepaspectratio=true,width=0.8\linewidth]{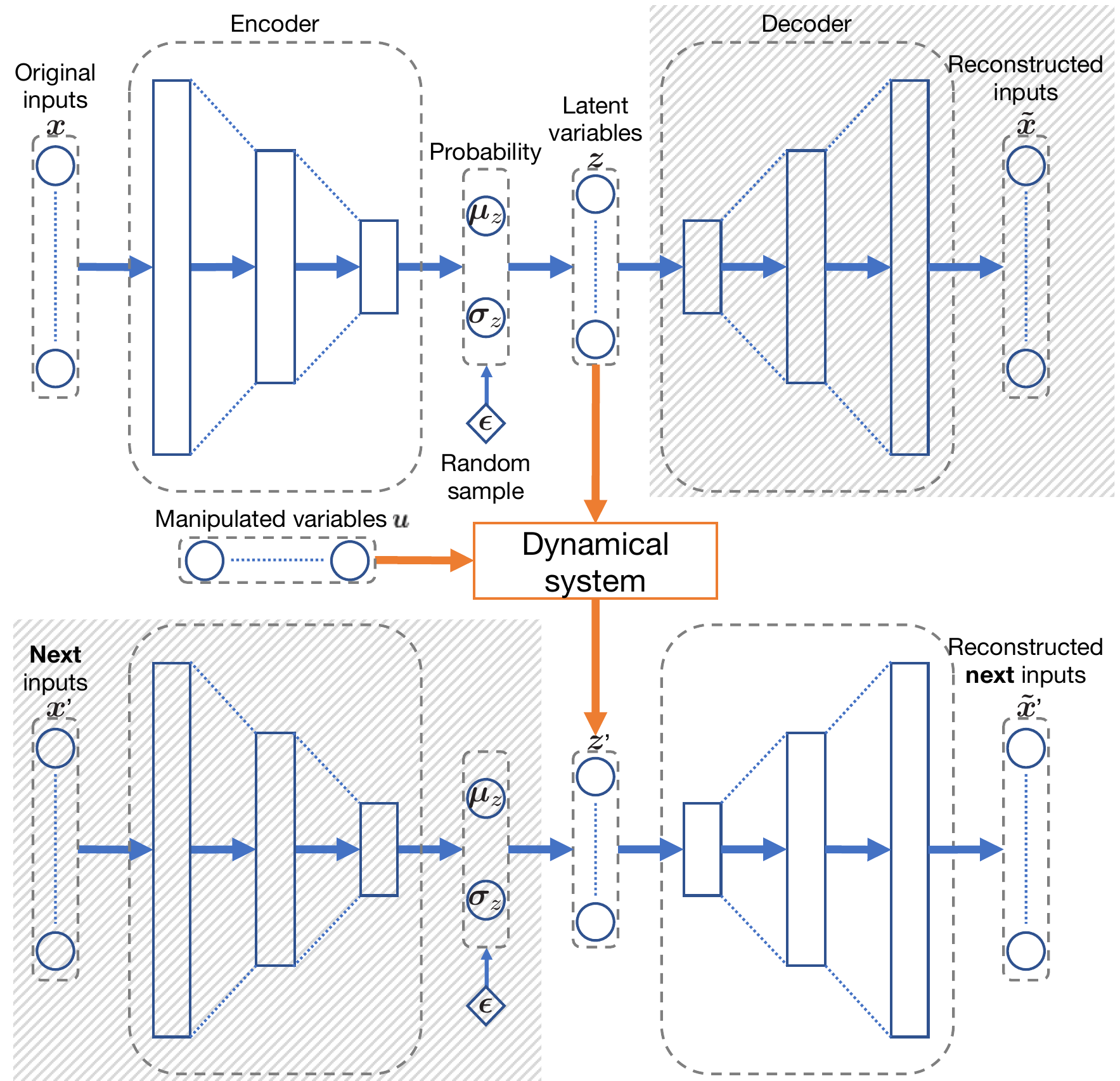}
    \caption{Dynamics model in latent space:
        upper and lower components are the same network structure as the standard VAE;
        the next inputs $\boldsymbol{x}^\prime$ is predicted according to the current inputs $\boldsymbol{x}$ and the given manipulated variables $\boldsymbol{u}$;
        in this paper, a dynamical system is given to be time-varying linear and diagonal.
    }
    \label{fig:structure_dynamics}
\end{figure}

Another advantage of Tsallis statistics is given by the $q$-likelihood defined in Eq.~\eqref{eq:q_likelihood}.
Its $q$-logarithm can be converted to the sum of the $q$-log likelihoods of respective samples without independency (i.e., with $q$-independency) between each sample.
In other words, even if the inputs are sampled from dynamical systems with a transition probability, the proposed q-VAE can be applied as is.

A simple dynamical system is designed in reference to the literature~\cite{watter2015embed,karl2017deep} (Fig.~\ref{fig:structure_dynamics}).
Here, a trajectory of inputs $X = \{\boldsymbol{x}_t\}_1^T$, where $T$ is the maximum time step, is generated from the following dynamics with parameters $\boldsymbol{\eta}$.
\begin{align}
    \boldsymbol{z}_{t} &\sim \rho(\boldsymbol{z}_{t} \mid \boldsymbol{x}_t; \boldsymbol{\phi})
    \\
    \boldsymbol{z}_{t+1} &\sim p(\boldsymbol{z}_{t+1} \mid \boldsymbol{z}_t, \boldsymbol{u}_t; \boldsymbol{\eta})
    \label{eq:dyn_latent}\\
    \boldsymbol{x}_{t+1} &\sim p(\boldsymbol{x}_{t+1} \mid \boldsymbol{z}_{t+1}; \boldsymbol{\theta})
\end{align}
where $\boldsymbol{u}$ denotes the manipulated variables of the dynamical system.

In that time, $\mathcal{L}_{(\beta, q)}(X)$ in Eq.~\eqref{eq:elbo_beta_qvae} is redefined as follows.
\begin{align}
    \mathcal{L}_{(\beta, q)}^\mathrm{dyn}(X) &= \sum_{t=1}^T
    \frac{\ln_q{p(\boldsymbol{x}_{t+1} \mid \boldsymbol{z}_{t+1}^{\boldsymbol{\eta}} ; \boldsymbol{\theta})}}{\beta_q(\boldsymbol{x}_t, \boldsymbol{z}_t)}
    \nonumber\\
    &- \beta \mathrm{KL}_q(\rho(\boldsymbol{z} \mid \boldsymbol{x}_t ; \boldsymbol{\phi}) \mid \mid p(\boldsymbol{z}))
    \label{eq:dyn_elbo}
\end{align}
where $\boldsymbol{z}_{t+1}^{\boldsymbol{\eta}}$ denotes $\boldsymbol{z}_{t+1}$ as predicted by the latent dynamics.
When maximizing this function, $\boldsymbol{\eta}$ must also be optimized implicitly.
However, the information about the dynamics would penetrate the decoder (and encoder), and $\boldsymbol{\eta}$ would fail to represent the latent dynamics.
Therefore, in addition to $\mathrm{KL}_q$, a further constraint is applied to the constrained optimization problem (Section~\ref{subsec:implementation}).
\begin{align}
    - \sum_{t=1}^T \ln \rho(\boldsymbol{z}_{t+1}^{\boldsymbol{\eta}} \mid \boldsymbol{x}_{t+1} ; \boldsymbol{\phi}) < \epsilon
    \nonumber
\end{align}
Similarly, this constraint can be rewritten as an additional maximization target with weight $\gamma$ as follows.
\begin{align}
    \mathcal{L}^\mathrm{latent}(X) = \gamma \sum_{t=1}^T \ln \rho(\boldsymbol{z}_{t+1}^{\boldsymbol{\eta}} \mid \boldsymbol{x}_{t+1} ; \boldsymbol{\phi})
    \label{eq:dyn_likelihood}
\end{align}
Such a constraint is also introduced in reference to the literature~\cite{watter2015embed}; however, this can be applied even if only the sampled latent variables are transited to the next ones using Eq.~\eqref{eq:dyn_latent}.

Note that the general form of the latent dynamics is simplified as much as possible to reduce computational cost.
The general (i.e., nonlinear) dynamics are regarded to be time-varying linear, as is used in general nonlinear control via first-order Taylor expansion.
As mentioned previously, the proposed q-VAE is suitable for disentangled representation; therefore, latent variables are ultimately independent of each other.
The latent dynamics are simplified as follows.
\begin{align}
    \boldsymbol{z}_{t+1} = \mathrm{diag}(\boldsymbol{a}(\boldsymbol{z}_t; \boldsymbol{\eta})) \boldsymbol{z}_t + B(\boldsymbol{z}_t; \boldsymbol{\eta}) \boldsymbol{u}_t
    \label{eq:linear_model}
\end{align}
where $\boldsymbol{a}$ and $B$ are the outputs of the network with inputs $\boldsymbol{z}$ and parameters $\boldsymbol{\eta}$.
In fact, this latent dynamics design is expected to cause modeling an error unless the latent variables are encoded to match this dynamics using disentangled representation learning.

\section{MNIST benchmark}

\begin{figure*}[tb]
    \centering
    \begin{minipage}[t]{0.32\linewidth}
        \centering
        \includegraphics[keepaspectratio=true,width=\linewidth]{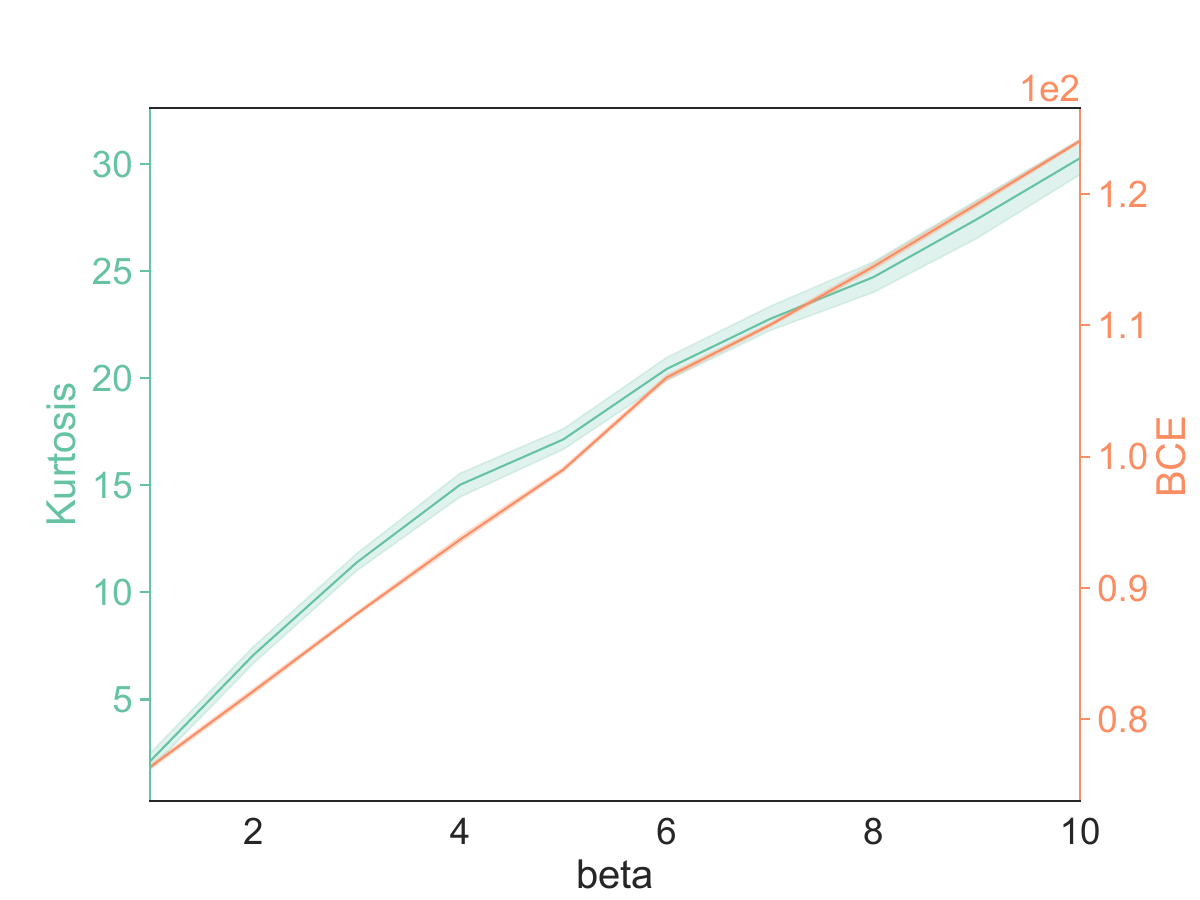}
        {\footnotesize (a) Effects of $\beta$}
    \end{minipage}
    \begin{minipage}[t]{0.32\linewidth}
        \centering
        \includegraphics[keepaspectratio=true,width=\linewidth]{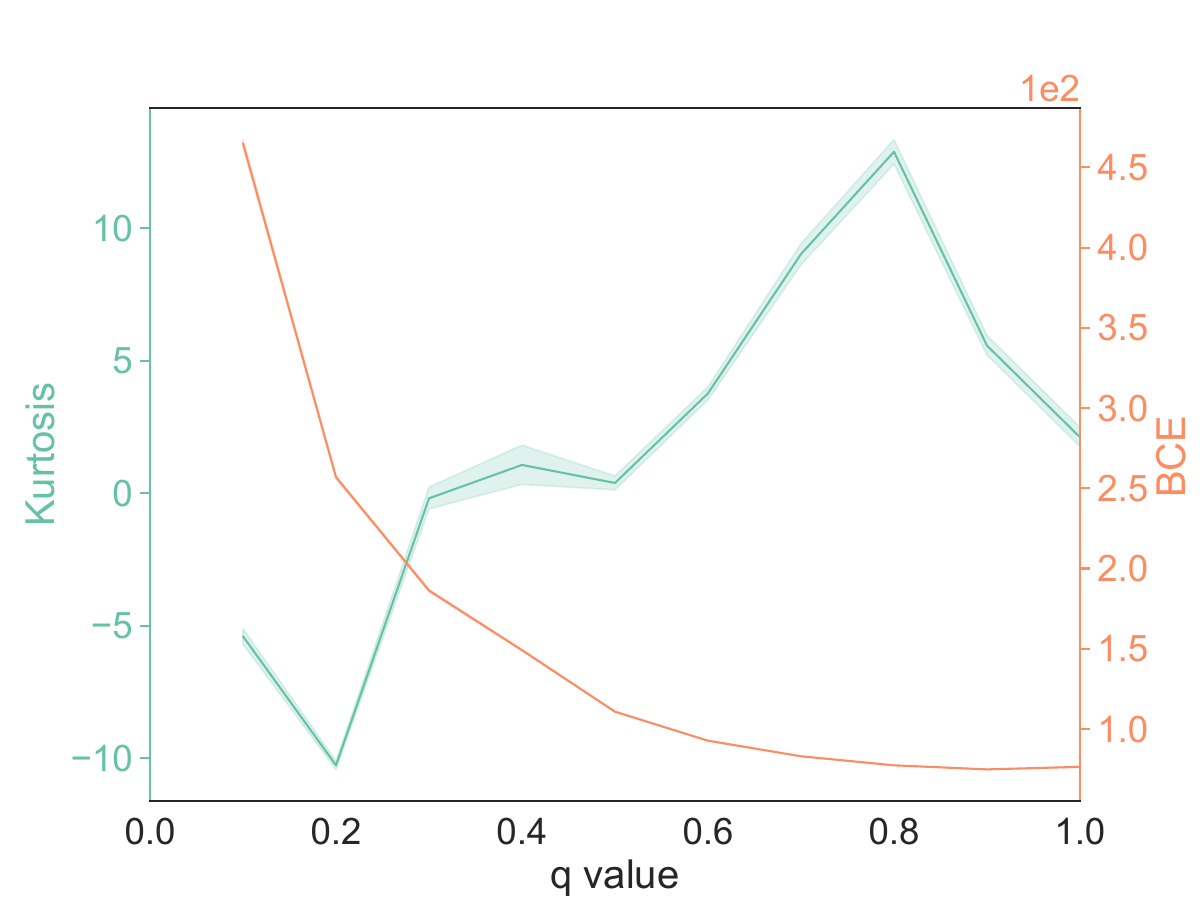}
        {\footnotesize (b) Effects of $q$ value}
    \end{minipage}
    \begin{minipage}[t]{0.32\linewidth}
        \centering
        \includegraphics[keepaspectratio=true,width=\linewidth]{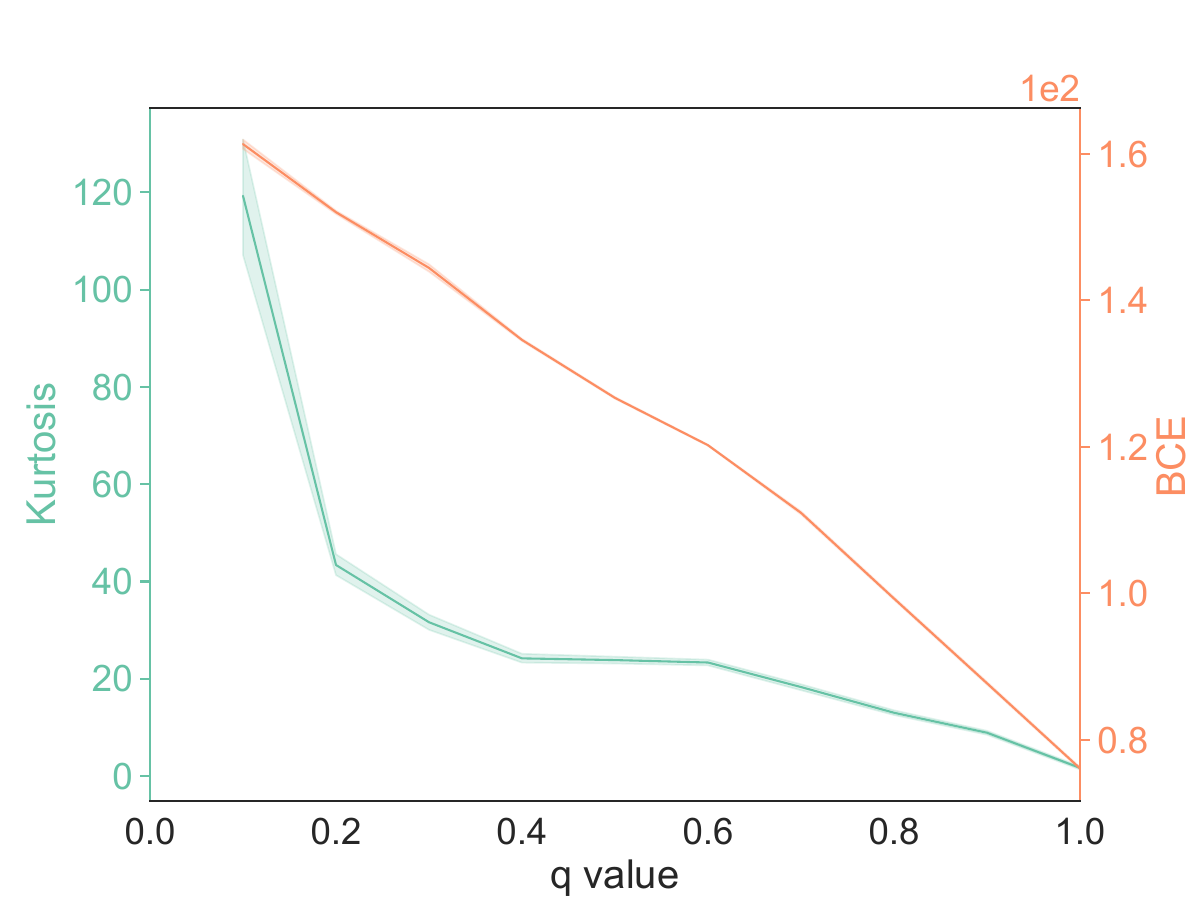}
        {\footnotesize (c) Effects of $q$ value on $\mathcal{L}_{(1, q)}^\mathrm{smp}$}
    \end{minipage}
    \caption{Criteria relative to hyperparameters.
        The shapes of the respective plots imply the qualitative effects of hyperparameters.
        (a) Kurtosis and BCE increase approximately linearly as $\beta$ increases.
        In other words, a tradeoff between the disentangled representation and reconstruction abilities is inevitable in $\beta$-VAE;
        (b) BCE is inversely proportional to $q$ value, while kurtosis became a unimodal-like shape.
        (c) The simplified q-VAE defined in Eq.~\eqref{eq:elbo_beta_qvae2} shows that kurtosis and BCE are inversely proportional to the value of $q$.
    }
    \label{fig:result_mnist_value}
\end{figure*}

\begin{figure}[tb]
    \centering
    \begin{minipage}[t]{0.49\linewidth}
        \centering
        \includegraphics[keepaspectratio=true,width=\linewidth]{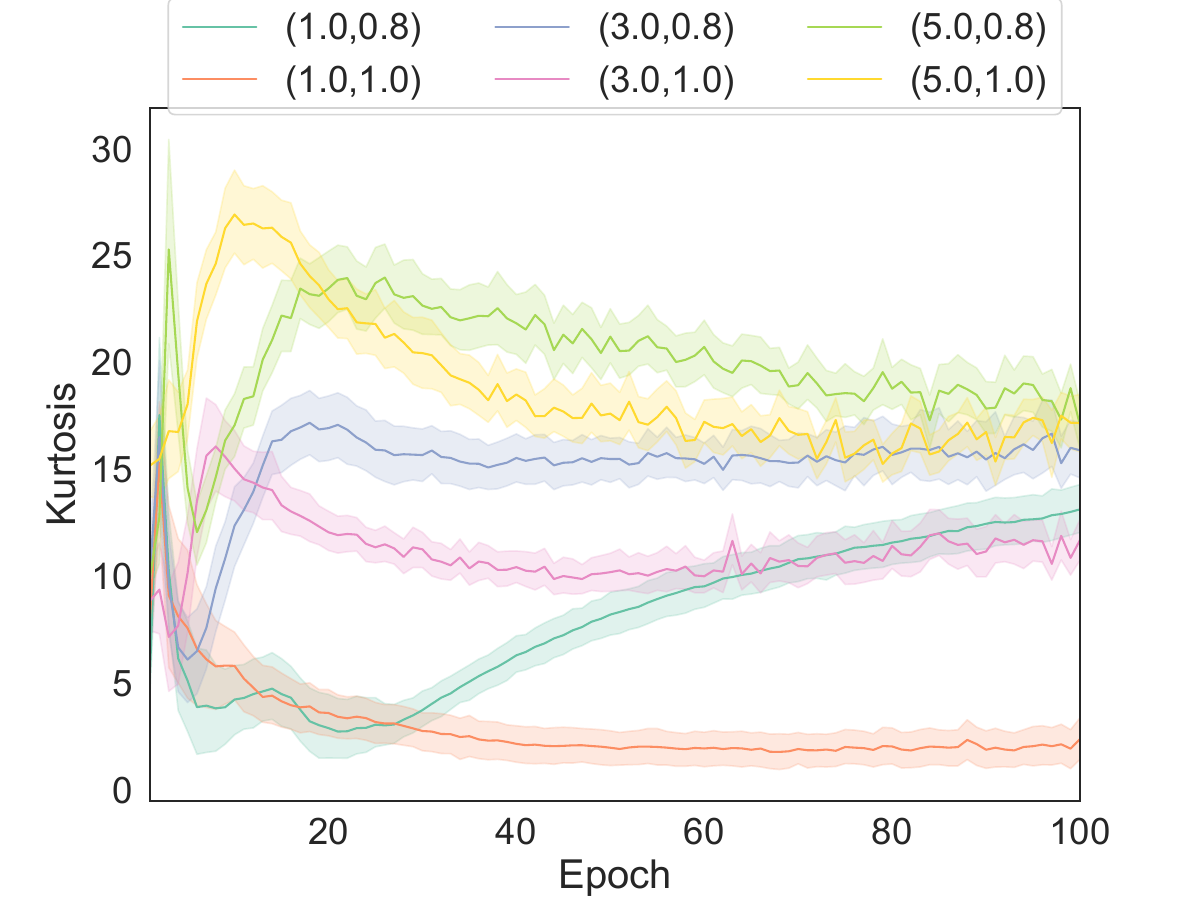}
        {\footnotesize (a) Kurtosis}
    \end{minipage}
    \begin{minipage}[t]{0.49\linewidth}
        \centering
        \includegraphics[keepaspectratio=true,width=\linewidth]{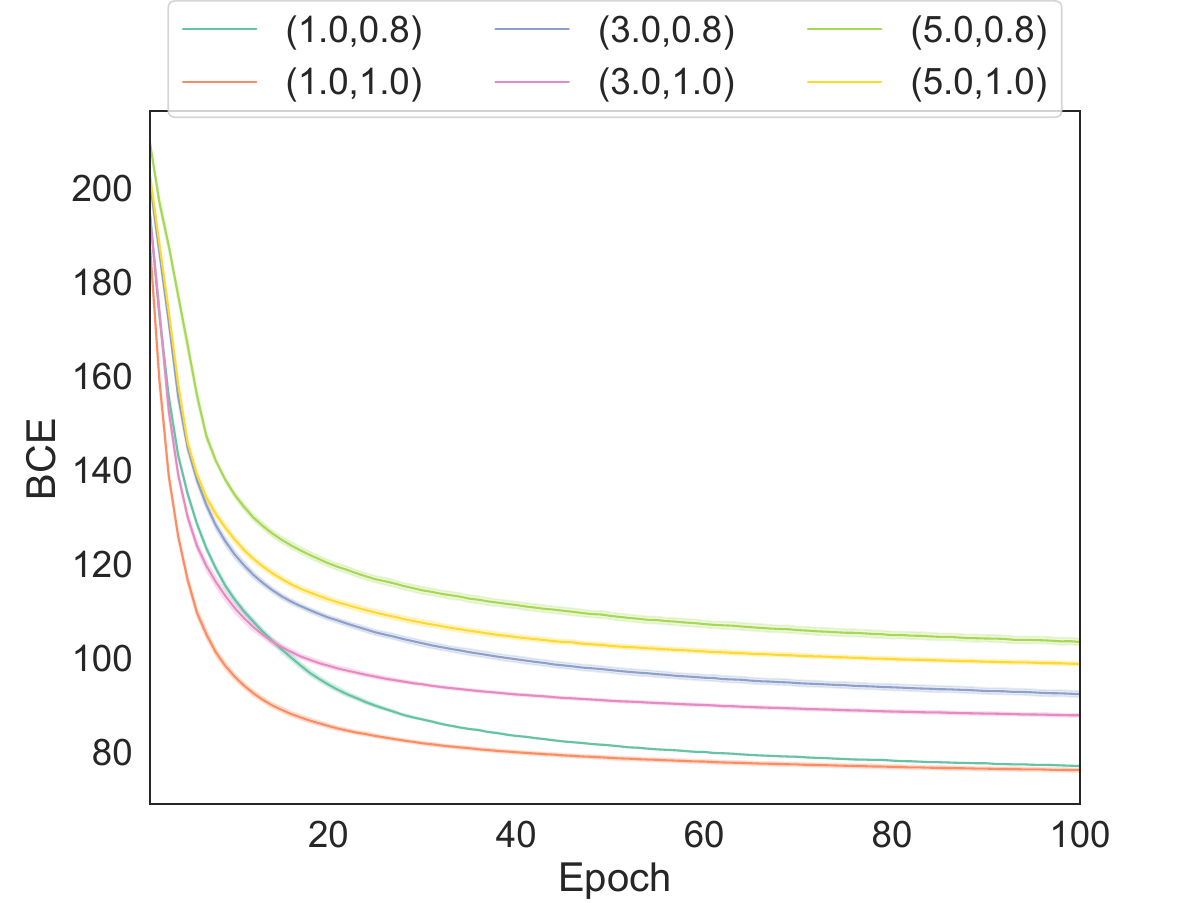}
        {\footnotesize (b) BCE}
    \end{minipage}
    \caption{Learning curves of the $(\beta, q)$-VAEs.
        The respective performances converge at 100 epoch.
        $\beta$-VAE effectively acquires large kurtosis at the expense of BCE.
        q-VAE increased kurtosis and suppressed the deterioration of BCE.
    }
    \label{fig:result_mnist_method}
\end{figure}

Here, we verify the performance of disentangled representation using the proposed q-VAE.
The MNIST dataset, which contains $28\times28=784$-dimensional images with handwritten numbers $0\sim9$, was used in this evaluation.

\subsection{Criteria of disentangled representation}

To evaluate how the latent space obtains a disentangled representation, we consider the fact that a disentangled representation attempts to gain independent axes with essential information.
This concept is closely related to independent component analysis (ICA)~\cite{girolami1996negentropy}.
Therefore, kurtosis in the latent space is employed according to ICA as a criterion of disentangled representation.

Specifically, kurtosis from the sampled data $Z = \{ \boldsymbol{z}_n\}_1^N$ where $\boldsymbol{z}_n \sim \rho(\boldsymbol{z} \mid \boldsymbol{x}_n ; \boldsymbol{\phi})$, $\kappa$, is defined as Mardia's kurtosis~\cite{mardia1970measures}.
\begin{align}
    \kappa = \frac{1}{N} \sum_{n=1}^N \left \{ (\boldsymbol{z}_n - \boldsymbol{\mu}_z)^\top \Sigma_z^{-1} (\boldsymbol{z}_n - \boldsymbol{\mu}_z) \right \} - d(d + 2)
\end{align}
where $\boldsymbol{\mu}_z$ and $\Sigma_z$ denote the mean and covariance of $Z$, respectively.
A greater $\kappa$ values indicates better disentangled representation.

In addition, reconstruction error criterion is important to demonstrate how much the infomative latent space is achieved.
Therefore, binary cross entropy (BCE) is employed.
Here, a smaller BCE value indicates better reconstruction.

Note that these two criteria are computed only from the test data.

\subsection{Network structure}

With the proposed q-VAE, the difference from the baselines is only related to ELBO (i.e., the loss function); therefore, the same network structure was employed for all compared methods.
In addition, the networks were constructed using PyTorch~\cite{paszke2017automatic}.

The images were converted to 784-dimensional vectors.
The encoder included five fully-connected layers: 500 neurons, 275 neurons, 50 neurons, and 20 neurons corresponding to $\boldsymbol{\mu}$ and $\boldsymbol{\sigma}$ of the posterior (i.e., $d = 10$).
The decoder also had five fully-connected layers: 255 neurons, 500 neurons, and 784 neurons corresponding to the reconstructed inputs.

In addition, the outputs from the intermediate layers of both the encoder and decoder were processed via layer normalization~\cite{ba2016layer} and Swish functions~\cite{ramachandran2017searching,elfwing2018sigmoid}.
We found that appropriate normalization techniques, e.g., layer normalization, are important relative to reduce the variance of kurtosis, and the activation function in the intermediate layers only improved the BCE value.

\subsection{Results}

The effects of $\beta$ in $\beta$-VAE~\cite{higgins2017beta} and $q$ value were investigated.
Here, (a) $\beta = [1, 10]$ with an increment of $1$ and (b) $q = [0.1, 1]$ an increment of $0.1$ were tested over 50 trials with 100 epochs. The results are plotted in Figs.~\ref{fig:result_mnist_value}(a) and (b), respectively.
Note that random seeds in the respective trials are given as the number of trials.

As expected, $\beta$-VAE increased the kurtosis (i.e., the disentangled representation ability) as $\beta$ increased (Fig.~\ref{fig:result_mnist_value}(a)).
However, the BCE valued deteriorated linearly along with $\beta$.
Here, $\beta$-VAE suffered a tradeoff problem between disentangled representation and reconstruction abilities, which may make the design of $\beta$ difficult.

In contrast, as shown in Fig.~\ref{fig:result_mnist_value}(b), the kurtosis has a peak at $q = 0.8$, and BCE was decreased monotonically as the value of $q$ increased.
In addition, BCEs in the $q > 0.7$ cases appeared to be sufficiently small, which suggests the value of $q$ should be approximately $0.8$ to achieve disentangled representation and effective reconstruction abilities.
In other words, the above different behaviors from $\beta$-VAE are achieved by the proposed q-VAE with a simple implementation, a slight increase in computational cost, and minimal effort to optimize the hyperparameters.
Here, one concern is that a small $q$ value makes Tsallis divergence $\mathrm{KL}_q$ small and reduces reconstruction accuracy by increasing $\beta_q$.

The results obtained with Eq.~\eqref{eq:elbo_beta_qvae2}, which was proposed to avoid such behavior, are shown in Fig.~\ref{fig:result_mnist_value}(c).
Due to the stronger and stable constraint to the prior, kurtosis increased extremely as $q$ decreased.
However, as expected, the stronger constraint caused the BCE to increase, even in $q \simeq 1$.
In other words, the same tradeoff as in $\beta$-VAE occurred; thus, the original derivation in Eq.~\eqref{eq:elbo_qvae} is more desirable compared to the simplified version in Eq.~\eqref{eq:elbo_beta_qvae2}.

Next, q-VAE was combined with $\beta$-VAE, as introduced in Eq.~\eqref{eq:elbo_beta_qvae}.
The learning curves of several combinations are shown in Fig.~\ref{fig:result_mnist_method}, where tuples denote $(\beta, q)$.
By focusing on pairs of i) the $(1.0, 0.8)$- and $(3.0, 1.0)$-VAEs, and ii) the $(3.0, 0.8)$- and $(5.0, 1.0)$-VAEs, we found that the proposed q-VAE increased the kurtosis as much as $\beta$-VAE while making BCE smaller than that of $\beta$-VAE.
In particular, the $(1.0, 0.8)$-VAE achieved superior disentangled representation ability than the $(3.0, 1.0)$-VAE, while also achieving similar reconstruction performance as the $(1.0, 1.0)$-VAE.

\section{Simulation to learn latent dynamics}

\subsection{Dataset}

\begin{figure}[tb]
    \centering
    \includegraphics[keepaspectratio=true,width=0.65\linewidth]{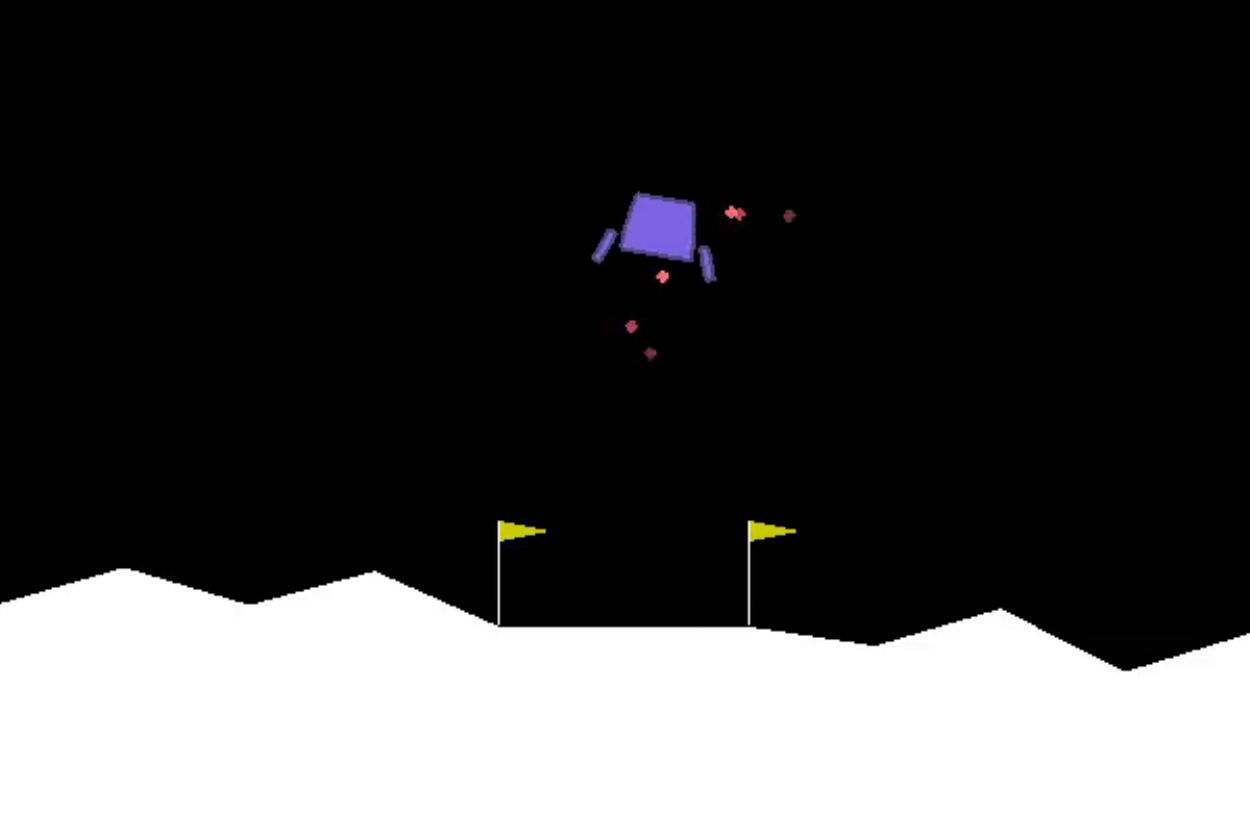}
    \caption{LunarLanderContinuous-v2~\cite{brockman2016openai}.
        A lander attempts to land safely inside two flags by controlling a main engine and two side engines.
        This problem has an eight-dimensional state and three-dimensional action spaces.
    }
    \label{fig:snap_lunar}
\end{figure}

As a proof of concept, a dynamical simulation provided by Open AI Gym~\cite{brockman2016openai}, i.e., LunarLanderContinuous-v2 (Fig.~\ref{fig:snap_lunar}), was learned by the proposed q-VAE using Eqs.~\eqref{eq:dyn_elbo} to \eqref{eq:linear_model}.
This simulation observes eight-dimensional state space data and is controlled via a three-dimensional action space.
Here, an expert controls the lander to land it safely in the permitted area via a keyboard interface.
In addition, the state and action pairs were collected as a non-i.i.d. expert's trajectory data.
In total, the dataset comprises 150 trajectories for training (and validation) and 50 trajectories for testing.
All tuples $(\boldsymbol{x}_t, \boldsymbol{u}_t, \boldsymbol{x}_{t+1})$ in the 150 trajectories were divided into 80 \% training data and 20 \% validation data.
Note that, during validation (after each training epoch), only the state prediction error was evaluated to investigate underfitting and overfitting.
After training, the compared methods attempted to predict the state (and latent variables) trajectories.

In this experiment, the prediction errors were the primary concern to demonstrate that the proposed q-VAE can extract latent dynamics, which is useful to reduce the computational cost of nonlinear model predictive control (MPC)~\cite{koenemann2015whole,amos2018differentiable}.
Therefore, the mean squared error (MSE) between the predicted and true states (or the predicted and encoded latent variables) is evaluated.
In addition, when applying the learned model to MPC, long-term prediction is more important; therefore, all states (and latent variables) were predicted from the initial state by repeatedly going through the latent dynamics.
Totally, four types of MSEs are given, i.e., $1$-step state, $1$-step latent, $T$-step state, and $T$-step latent, and these were computed using only the test data with 50 trajectories.
Note that each trajectory has approximately 500 steps; thus, the proposed q-VAE had to predict states from one to approximately 500 steps future only using the initial state and actions at the respective times.

\subsection{Network structure}

Due to real number inputs, all network layers were fully connected.
In this study, three network structures were prepared to demonstrate the network-invariant performance.
The numbers of layer neurons are listed in Table~\ref{tab:network}. As can be seen, V1 is the largest structure, V2 is a moderate encoder with the smallest latent dynamics, and V3 is the simplest design.
Note that a decoder inverted the hidden layers in the encoder with different parameters.
In addition to the MNIST benchmark, layer normalization~\cite{ba2016layer} and Swish functions~\cite{ramachandran2017searching,elfwing2018sigmoid} were employed in this evaluation.

\begin{table}[tb]
    \caption{Network designs for learning latent dynamics}
    \label{tab:network}
    \centering
    {
    \begin{tabular}{ccc}
        \hline\hline
        Version & Encoder & Latent dynamics
        \\
        \hline
        V1 & [500, 400, 300, 200, 100] & [100, 100, 100, 100, 100]
        \\
        V2 & [250, 200, 150, 100] & [50, 50, 50]
        \\
        V3 & [100, 100, 100] & [100, 100, 100]
        \\
        \hline\hline
    \end{tabular}
    }
\end{table}

\subsection{Results}

\begin{figure*}[tb]
    \centering
    \begin{minipage}[t]{0.32\linewidth}
        \centering
        \includegraphics[keepaspectratio=true,width=\linewidth]{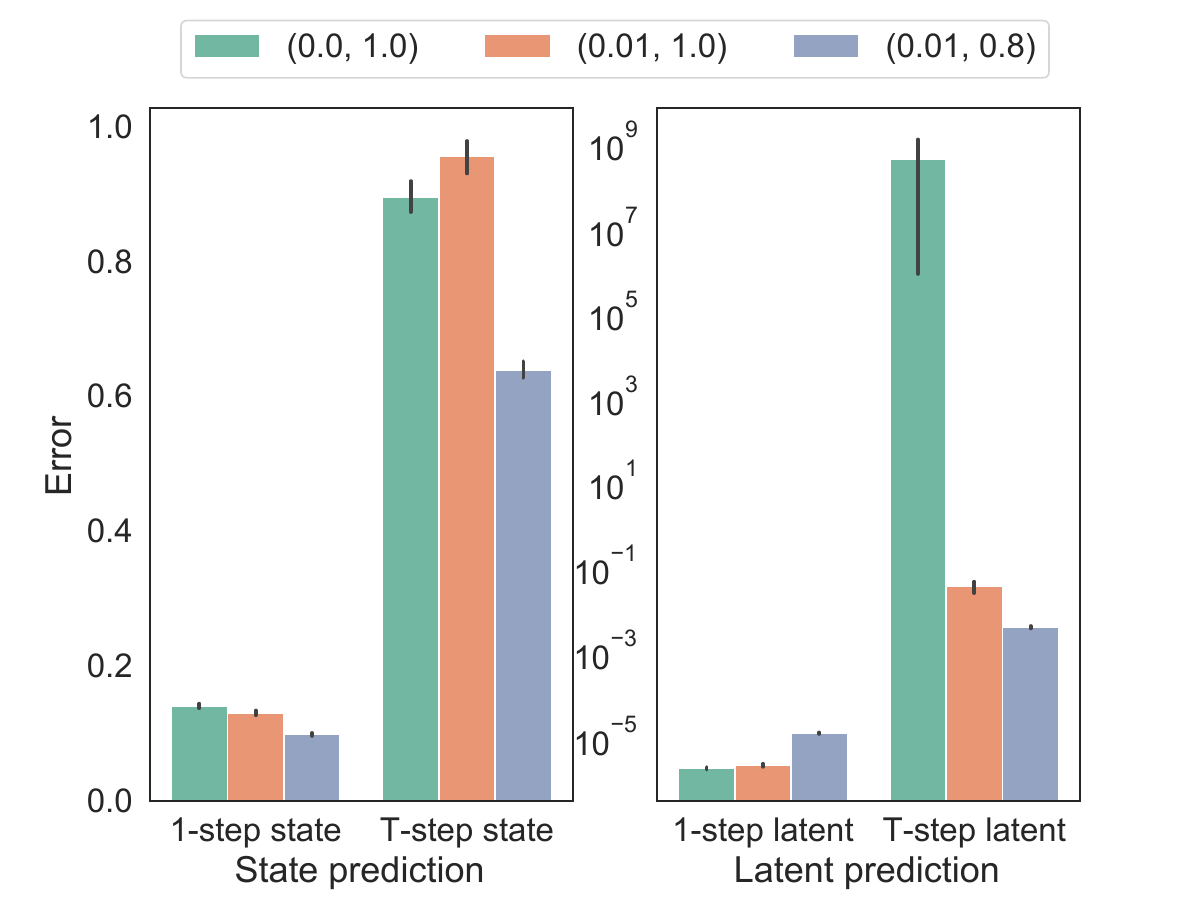}
        {\footnotesize (a) V1}
    \end{minipage}
    \begin{minipage}[t]{0.32\linewidth}
        \centering
        \includegraphics[keepaspectratio=true,width=\linewidth]{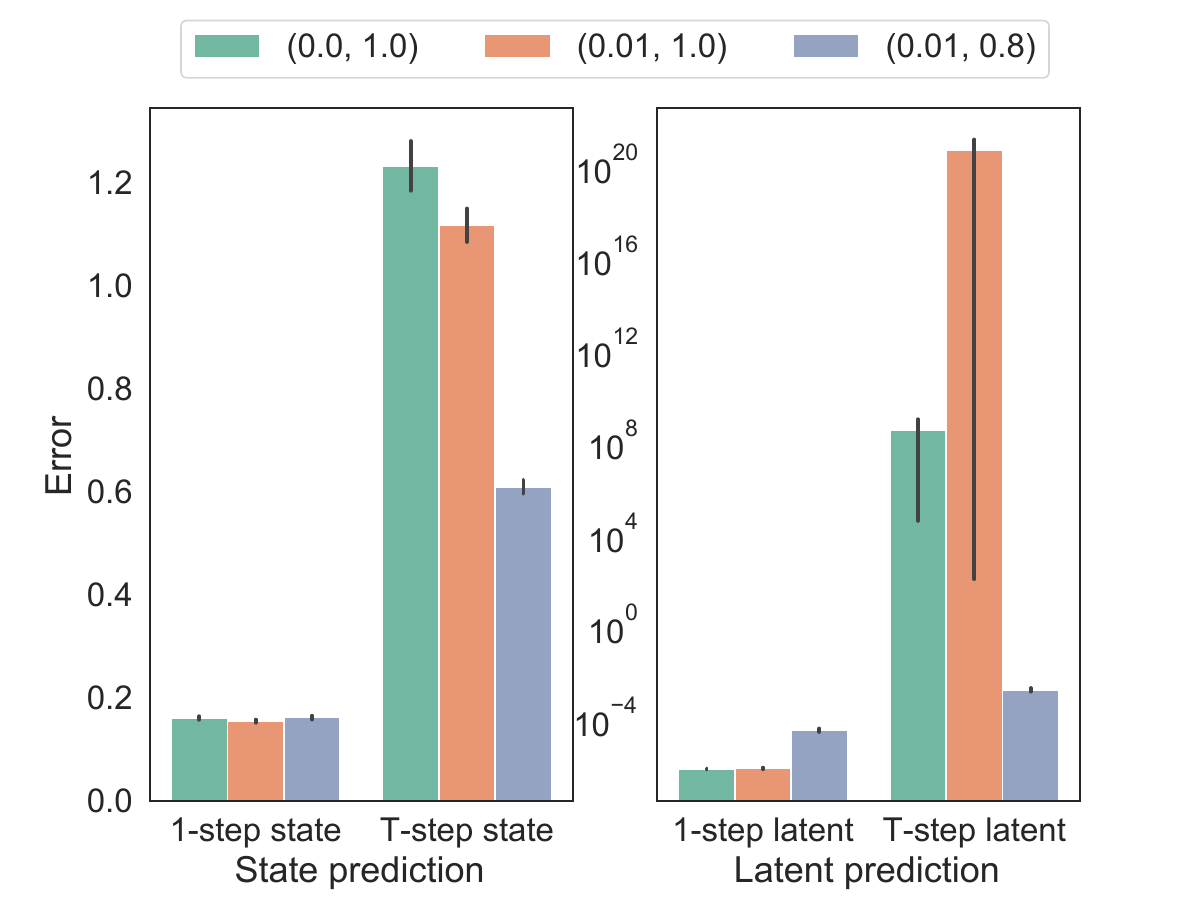}
        {\footnotesize (b) V2}
    \end{minipage}
    \begin{minipage}[t]{0.32\linewidth}
        \centering
        \includegraphics[keepaspectratio=true,width=\linewidth]{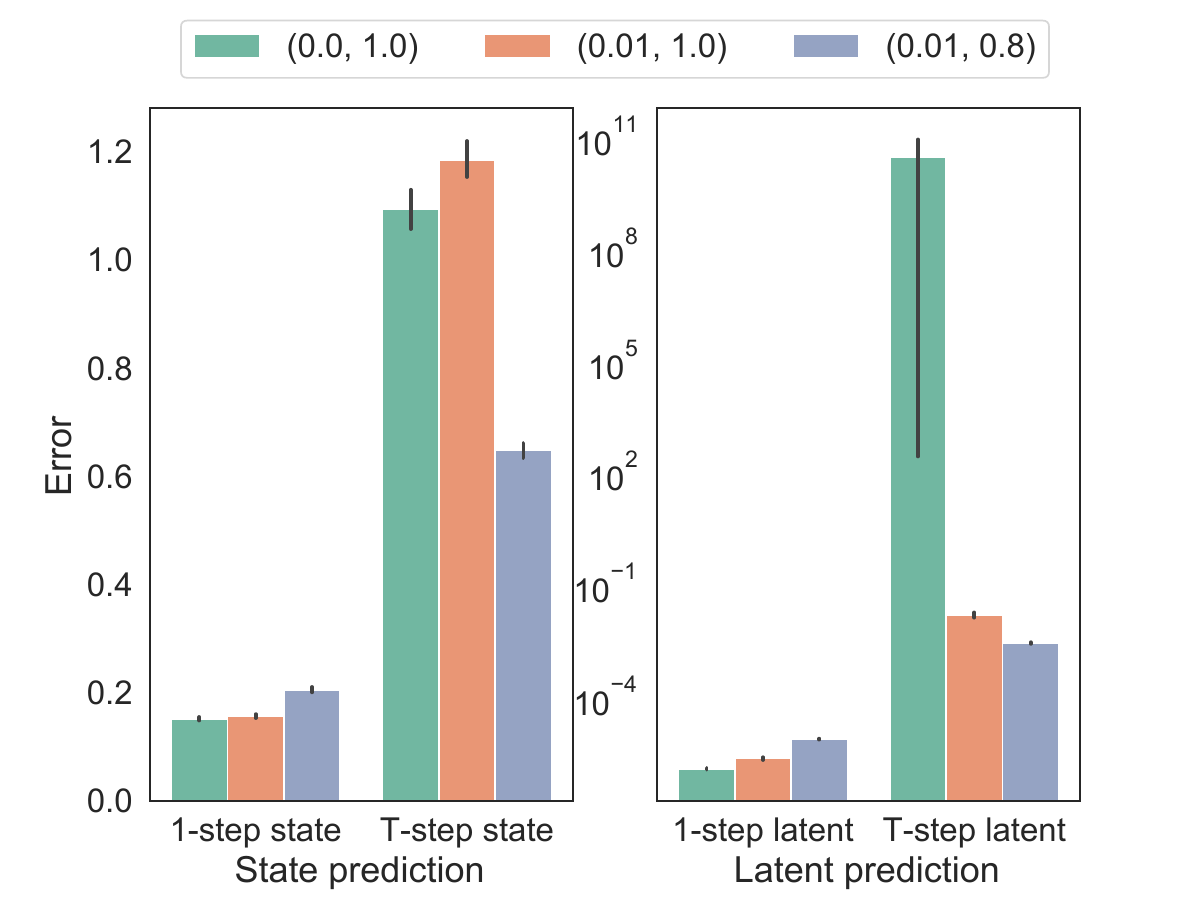}
        {\footnotesize (c) V3}
    \end{minipage}
    \caption{Prediction errors of respective models.
        All methods succeeded in $1$-step prediction with the same accuracy.
        With all models, the proposed q-VAE gained better $T$-step prediction than the compared methods.
        In particular, the compared methods made the $T$-step latent diverge occasionally; however, this trend did not occur with the proposed q-VAE.
    }
    \label{fig:result_lunar_score}
\end{figure*}

Three conditions with different $(\beta, q)$ were compared: $(0.0, 1.0)$ to demonstrate that the standard autoencoder does not have continuity in the latent space, $(0.01, 1.0)$ as a baseline, and $(0.01, 0.8)$ as the proposed q-VAE.
Here, $\beta$ was less than the MNIST case because the ratio between the input and latent dimensions was approximately 100 times different, and the $\beta=1$ case demonstrated relatively strong regularization.
As common conditions, a three dimensional latent space was given, and $\gamma$ in Eq.~\eqref{eq:dyn_likelihood} was set to $0.1$.
For each condition, 50 trials with different random seeds were conducted, and the mean of the prediction error during each trajectory was computed for the test data.

The results are summarized in Fig.~\ref{fig:result_lunar_score}.
Regardless of the conditions and network structures, ability to predict the $1$-step state and latent variables was demonstrated.
In contrast, $T$-step prediction demonstrated the superiority of q-VAE.
For $T$-step latent variable prediction, the proposed q-VAE method yielded stable results, although the compared methods made it diverge occasionally.
This implies that the latent space extracted by the proposed q-VAE provides a natural representation of dynamics.
As a result, $T$-step state prediction was improved by q-VAE.
Interestingly, although the other methods with the largest network structure (i.e., V1) gained better results compared to the other structures, the proposed q-VAE achieved the same excellent prediction performance regardless of network structure.

\section{Real robot experiment}

\subsection{Conditions}

\begin{figure}[tb]
    \centering
    \includegraphics[keepaspectratio=true,width=0.65\linewidth]{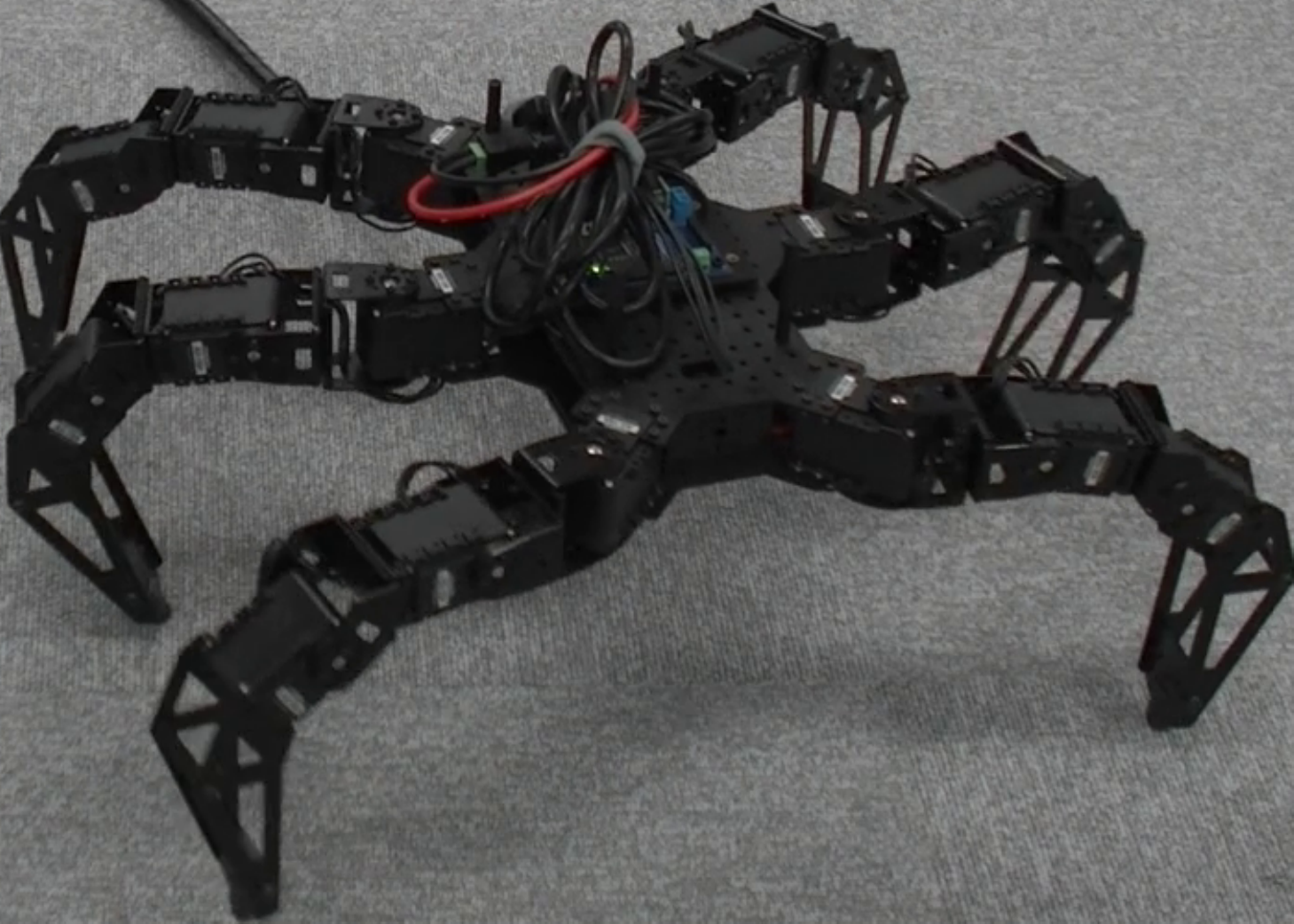}
    \caption{ PhantomX hexapod developed by Trossen Robotics.
        Each leg is controlled by inverse kinematics toward its reference given as an oscillator's phase.
    }
    \label{fig:snap_px}
\end{figure}

In this experiment, the walking motion of a hexapod robot (Trossen Robotics; Fig.~\ref{fig:snap_px}) was evaluated to learn its dynamics.
The hexapod robot uses central pattern generators (CPGs)~\cite{lewis1993genetic,owaki2017minimal} to generate periodic walking.
The dynamics of six oscillators corresponding to the robot's legs are given as follows.
\begin{align}
    \dot{\boldsymbol{\xi}} = e^{-k\boldsymbol{e}} \omega + \boldsymbol{a}(\boldsymbol{\xi})
\end{align}
where $\boldsymbol{\xi}$ denotes the phases of the oscillators with natural frequency $\omega$, which is slowed according to the position errors of the legs $\boldsymbol{e}$ and gain $k$.
Note that this error term imitates the tegotae-based ~\cite{owaki2017minimal}.
$\boldsymbol{a}$ are computed according to the oscillator network, which attracts walking gait to tripod.

Here, the observed and reference joint angles are given as inputs.
In total, this experiment considered 36-dimensional state and six-dimensional action spaces.
Similar to the previous experiment, the dataset comprised 150 trajectories for training (and validation) and 50 trajectories for testing, and each trajectory involved 500 steps.
In addition, the criteria and hyperparameters used in the previous experiment were considered in this experiment.
Among the three network structures detailed in Table~\ref{tab:network}, V3 was employed in this experiment.
In addition, due to noisy observations, a robust optimizer~\cite{ilboudo2020tadam} was also employed.

\subsection{Results}

\begin{figure}[tb]
    \centering
    \begin{minipage}[t]{0.49\linewidth}
        \centering
        \includegraphics[keepaspectratio=true,width=\linewidth]{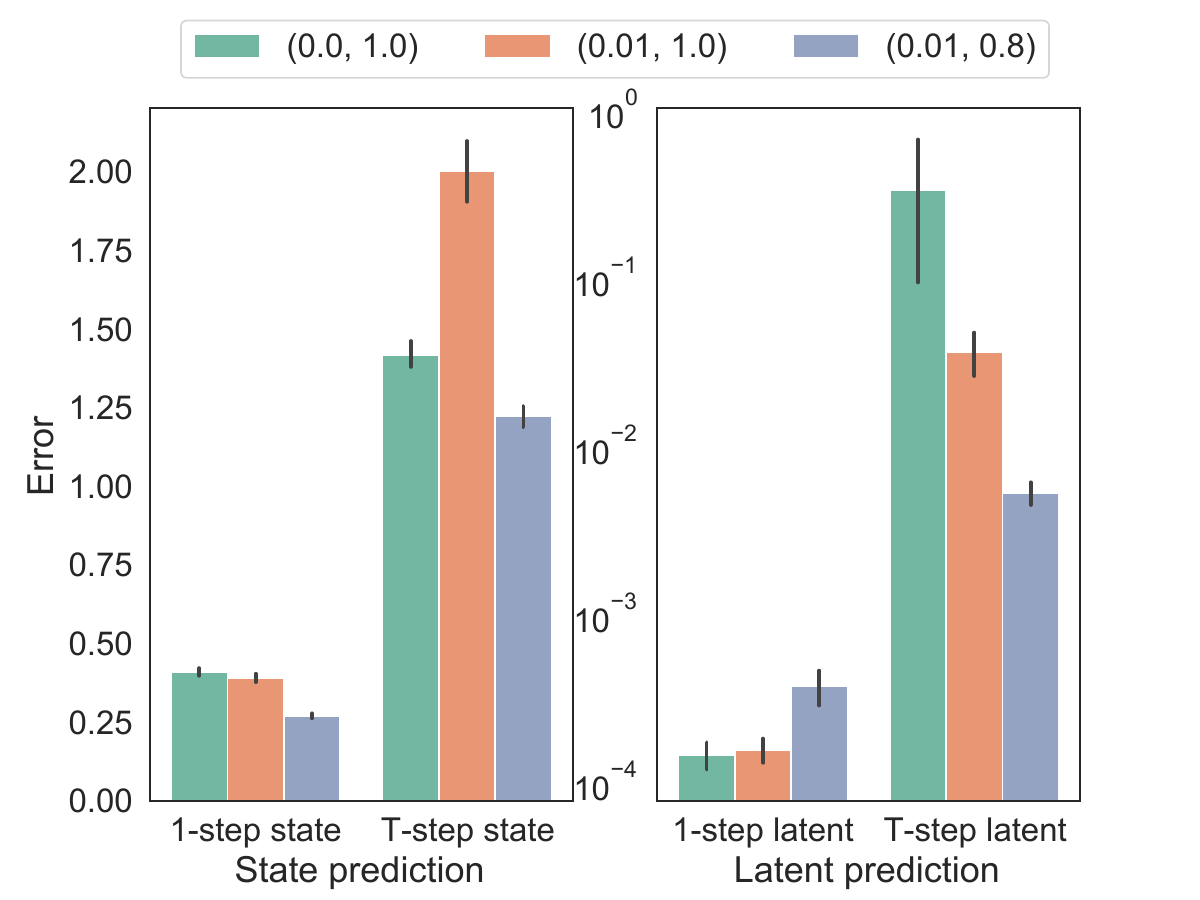}
        {\footnotesize (a) Prediction errors}
    \end{minipage}
    \begin{minipage}[t]{0.49\linewidth}
        \centering
        \includegraphics[keepaspectratio=true,width=\linewidth]{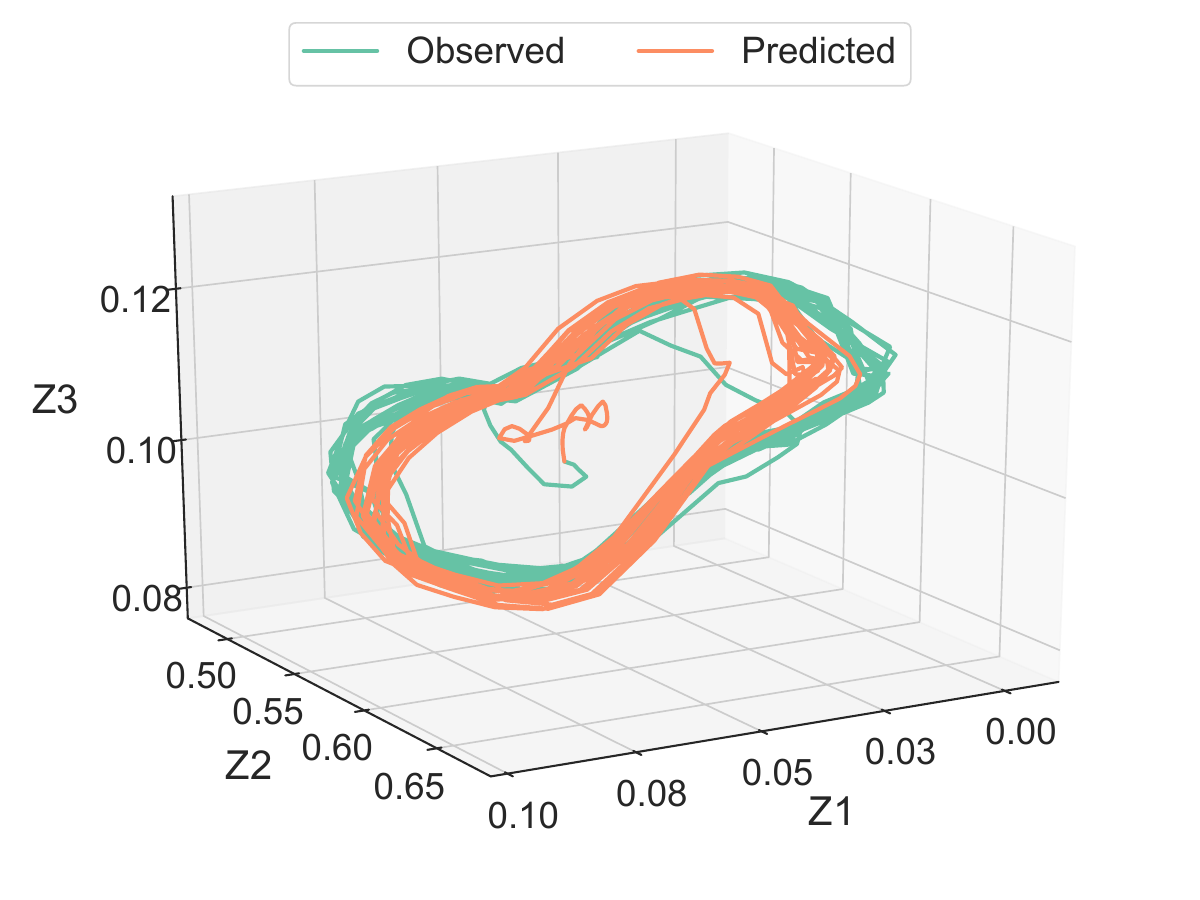}
        {\footnotesize (b) Trajectories in latent space}
    \end{minipage}
    \caption{Learning results of $(\beta, q)$-VAEs.
        (a) The proposed q-VAE improved prediction performance.
        (b) The trajectory predicted by the proposed q-VAE was attracted to almost the same periodic attractor as the observed attractor.
    }
    \label{fig:result_px}
\end{figure}

The results are shown in Fig.~\ref{fig:result_px}.
As can be seen, the proposed q-VAE improved prediction performance compared to the baselines.
As an example, prediction from the initial state and actions are visualized in the attached video.
The latent variables are shown in the right of Fig.~\ref{fig:result_px}.
As can be seen, the trajectory of the predicted latent variables were attracted to nearly the same periodic attractor as the observed ones.
In other words, the proposed q-VAE revealed the natural latent dynamics corresponding to the oscillator dynamics from real observation data.

\section{Conclusion}

In this paper, we have proposed the q-VAE method, which is derived according to Tsallis statistics .
Due to Tsallis statistics, the proposed q-VAE has three primary beneficial features, i.e., input-dependent $\beta_q$ according to the amount of information in the latent space, it demonstrates Tsallis divergence regularization rather than the standard KL divergence, and it relaxes the assumption of i.i.d. input data.
The first two features are suitable for disentangled representation learning, and the proposed q-VAE outperformed the baseline $\beta$-VAE approach on the MNIST benchmark.
In addition, the second feature, i.e., Tsallis divergence, was verified by testing a simplified version of q-VAE.
The third feature allows the proposed q-VAE to be used to learn latent dynamics from non i.i.d. data.
As a proof of concept, the proposed q-VAE was demonstrated to stably and accurately predicting future states (approximately 500 steps) from only the initial state and the action sequence.

Although the proposed q-VAE outperformed the compared baselines, it has several practical approximations.
In future, we expect that further improvement can be obtained by removing these approximations.
In addition, the proposed q-VAE was derived as a new base of VAE variants; thus, it can be integrated with the latest disentangled representation learning methods~\cite{burgess2018understanding,chen2018isolating,kim2018disentangling} to improve their performance.
Finally, the proposed q-VAE can be applied to real complex systems with high-dimensional input, e.g., vision systems, to control them relative to the prediction of future states in real time using the latest control theory.

%
%
%
\section*{ACKNOWLEDGMENT}

This work was supported by JSPS KAKENHI, Grant-in-Aid for Scientific Research (B), Grant Number 20H04265.

\bibliographystyle{IEEEtran}
{
\bibliography{qvae}
}

\end{document}